\documentclass[runningheads]{llncs}
\usepackage{graphicx}
\usepackage{url,hyperref,lineno,microtype,subcaption}
\usepackage[onehalfspacing]{setspace}
\usepackage{multirow}
\usepackage{amsmath}
\usepackage{geometry}
\begin{document}
\title{Fast and Accurate Object Detection on Asymmetrical \\Receptive Field}
%
%

\author{Tianhao Lin}
%
%
\institute{Technical Universsity of Munich, Garching, Germany
}

\maketitle              
\begin{abstract}
Object detection has been used in a wide range of industries. For example, in autonomous driving, the task of object detection is to accurately and efficiently identify and locate a large number of predefined classes of object instances (vehicles, pedestrians, traffic signs, etc.) from videos of roads. In robotics, the industry robot needs to recognize specific machine elements. In the security field, the camera should accurately recognize each face of people. With the wide application of deep learning, the accuracy and efficiency of object detection have been greatly improved, but object detection based on deep learning still faces challenges. Different applications of object detection have different requirements, including highly accurate detection, multi-category object detection, real-time detection, robustness to occlusions, etc. To address the above challenges, based on extensive literature research, this paper analyzes methods for improving and optimizing mainstream object detection algorithms from the perspective of evolution of one-stage and two-stage object detection algorithms. Furthermore, this article proposes methods for improving object detection accuracy from the perspective of changing receptive fields. The new model is based on the original YOLOv5 (You Look Only Once) with some modifications. The structure of the head part of YOLOv5 is modified by adding asymmetrical pooling layers. As a result, the accuracy of the algorithm is improved while ensuring the speed. The performances of the new model in this article are compared with original YOLOv5 model and analyzed from several parameters. And the evaluation of the new model is presented in four situations. Moreover, the summary and outlooks are made on the problems to be solved and the research directions in the future.

\end{abstract}
\section{Introduction}

In recent years, object detection has always been a fundamental problem in computer vision. Object detection can be divided into two major schools of thought due to different tendencies for effectiveness. One is two-stage object detection, which focuses more on accuracy, and the other is one-stage object detection, which focuses more on speed. Two-stage object detection, as the name implies, solves the problem in two stages. The first stage is the generation of the regions of interest(RoI) which is called Region Proposal and the extraction of features using convolutional neural networks. The second stage is to put the output of the first stage into the support vector machine (SVM) or CNN-based classifier to classify objects and then correct the objects’ positions using the bounding box regression. Two-stage object detection originated from Regions with CNN features (R-CNN) (\cite{girshick2014rich}). R-CNN uses a heuristic (Selective search) to reduce the redundancy of information and improve the speed of detection by firstly forming Region proposal before detection. In addition, the robustness of feature extraction is improved. The researchers then proposed a new neural network by applying a technique named Spatial Pyramid Pooling (SPP) (\cite{he2015spatial}), which not only reduces computational redundancy, but more importantly, breaks the bound of fixed-size input of fully collected layer. After SPP Net, Fast R-CNN (\cite{girshick2015fast}) emerged. Compared with the original Slow R-CNN, it has been optimized for speed. It changes the original serial structure to parallel structure. The algorithm performs regression on bounding box (Bbox) while classifying. But this was not good enough, so the researchers proposed Faster R-CNN (\cite{ren2015faster}). Different from the previous heuristic algorithm to produce region proposals, Faster R-CNN proposes a concept of Region Proposal Networks(RPN), which use neural networks to learn to generate region proposals. Meanwhile, the concept of anchor has been introduced in RPNs. The object detection of R-CNN series has been improved and evolved step by step to get the final Faster R-CNN algorithm, which has a great improvement in both accuracy and speed. However, there is still no way to achieve real-time object detection, so the One-Stage object detection algorithm was proposed later. 

One-stage object detection is a one-shot solution that directly regresses on the predicted object of interest. Compared to two-stage object detection, it is very fast and finds a balance between fast and accurate. 'You only look once' (YOLO) (\cite{redmon2016you}) is one of the representative algorithms. YOLO first resizes the image to a fixed size, then passes it through a set of convolutional neural networks, and finally connects it to fully convolutional layer to output the result directly, which is the basic structure of the whole network. The latest algorithms YOLOv5 has been able to obtain relatively satisfactory results. It's very fast while ensuring sufficient accuracy. Throughout the neural network model, we believe that the final feature map plays a critical role in the results. Each pixel in feature map has a corresponding receptive field (\cite{luo2016understanding}). The depth of feature map is deeper, the receptive field is larger. In the end of the neural network, YOLOv5 algorithm generates three different sizes of feature map. All of those pixels from these three feature maps have the same shape of receptive field——square. Moreover, for better detection of different shapes objects, each feature map has three different shapes of anchors. Now we have a new conjecture that if we change the shape of receptive field, the detection capability of the algorithm will be improved, and it will be easier to detect objects having different shapes. The feature map whose pixels have square receptive field can detect square object more easily. Conversely, the feature map whose pixels have rectangular receptive field can detect rectangular object more easily. Based on this conjecture, we make some modifications to the YOLOv5 model so that we can change the receptive fields of the final feature maps.

\section{Related Work}

In this section, we firstly introduce the development of YOLO (You Look Only Once). Then we introduce COCO (Common Objects in Context) dataset (\cite{lin2014microsoft}). Finally, we explain the metrics in the evaluation of YOLO algorithm.

\subsection{Development of YOLO}
As the pioneer of one-stage algorithm, YOLOv1 (\cite{redmon2016you}) was a big hit with its simple network structure and real-time detection speed on GPU, breaking the "monopoly" of R-CNN series and bringing a huge change to the field of object detection. YOLOv1 has many drawbacks when viewed from today's perspective, but back then, YOLOv1 was very popular and provided the framework basis for many one-stage algorithms later. The most important feature of YOLOv1 is that it uses only one convolutional neural network to achieve the purpose of object detection end-to-end. At CVPR 2016, following the YOLOv1 work, the original authors reintroduced YOLOv2 (or YOLO9000) (\cite{redmon2017yolo9000}). Compared with YOLOv1, YOLOv2 
introduced the anchor box mechanism proposed by the Faster R-CNN, the use of K-means clustering algorithm to obtain a better anchor box. The regression method of the bounding box was also adjusted. Later, YOLOv3 (\cite{redmon2018yolov3}) is proposed. The changes of YOLOv3 is not only using a better backbone network: DarkNet-53, but also using Feature Pyramid Networks (FPN) (\cite{lin2017feature}) technology and multi-level detection methods. YOLOv4 (\cite{bochkovskiy2020yolov4}) was proposed in April 2020. It achieves 43.5\% AP accuracy and 65 FPS on MS COCO dataset, a 10\% and 12\% improvement compared to YOLOv3, respectively. YOLOv4 uses the CSPDarkNet-53 network as a new backbone network, which had excellent speed and accuracy at the time. Shortly after YOLOv4 was proposed, Ultralytics came up with YOLOv5. YOLOv5 has no particular changes in network structure, but it has better performance on speed and accuracy.

\subsection{COCO Dataset}
The COCO dataset (\cite{lin2014microsoft}) is a large-scale dataset that can be used for image detection, semantic segmentation and image captioning. It has more than 330K images (220K of them are annotated), containing 1.5 million objects, 80 object categories (pedestrian, car, bicycle, etc.), 91 stuff categories (grass, wall, sky, etc.), five image descriptions per image, and 250K pedestrians with key point annotation. As for object detection, we use COCO2017. The number of training images are 118K, the number of validation images are 5K and there are 40K test images. Each label of images has 5 parameters, which are category, $x$ coordinate of centroid, $y$ coordinate of centroid, width $w$, and height $h$. The dataset contains 80 categories covering a large number of real-life scenarios, such as traffic, interviews, dances, animals, etc. These objects differ in scale, occlusion, pose, expression, and lighting conditions. Therefore, the training data is large enough to be a challenge for the detector.

\begin{table}[h]
    \centering
    \begin{tabular}{|c|c|c|c|c|}
    \hline & Training set & Validation set & Testing set & Total \\
    \hline Nr. of images & 118,287 & 5,000 & 40,670 & 163,957 \\
    \hline Percentage & $70 \%$ & $5 \%$ & $25 \%$ & $100 \%$ \\
    \hline
    \end{tabular}
    \caption{Basic statistics of the COCO dataset.}
    \label{tab:coco}
\end{table}

\subsection{Metrics}
Several metrics are widely used to evaluate the performance of object detection, which mainly include Precision-Recall (PR) curves and Average Precision (AP) (\cite{powers2020evaluation}). Before we explain this two metrics, we need first introduce the confusion matrix. 

\begin{table}[htpb]
    \centering
    \begin{tabular}{|c|c|c|}
    \hline \multirow{2}{*}{ Actual condition } & \multicolumn{2}{|c|}{ Predicted condition } \\
    \cline { 2 - 3 } & Positive & Negative \\
    \hline Positive & TP (True Positive) & $\mathrm{FN}$ (False Negative) \\
    \hline Negative & $\mathrm{FP}$ (False Positive) & $\mathrm{TN}$ (True Negative) \\
    \hline
    \end{tabular}
    \caption{Confusion table for classification results.}
    \label{tab:metric}
\end{table}

True Positive (TP) means that the sample is actually positive and the network also predicts the sample as positive. True Negative (TN) means that the sample is actually negative and the network also predicts the sample as negative. Therefore, if the result is TP or TN, the network makes true predictions. False Positive (FP) means the prediction is wrong, because the sample is actually negative but the network predicts it as positive. False Negative: the sample is actually positive but the network predicts it as negative, so this prediction is also wrong. Precision and Recall is a common pair of performance metrics based on confusion matrix. 

\begin{equation}
\text { Precision }=\frac{T P}{T P+F P}
\end{equation}

\begin{equation}
\text { Recall }=\frac{T P}{T P+F N}
\end{equation}

Average Precision (AP) represents the area under the Precision-Recall curve. Generally, the higher the value of AP, the better the performance of the classifier. The value of AP lies in [0,1]. A perfect classifier will have an AP value of 1. Each class has a AP (\cite{boyd2013area}). The mean Average Precision (mAP) is calculated by finding AP for each class and then average over a number of classes.

\begin{equation}
\mathrm{mAP}=\frac{1}{N} \sum_{i=1}^N \mathrm{AP}_i
\end{equation}

\noindent The mAP incorporates the trade-off between precision and recall and considers both false positives (FP) and false negatives (FN). This property makes mAP a suitable metric for most detection applications.

\section{Proposed Methodology}

\subsection{Architecture of YOLOv5}
\begin{figure}[htpb]
  \centering
  \includegraphics[width=0.7\textwidth]{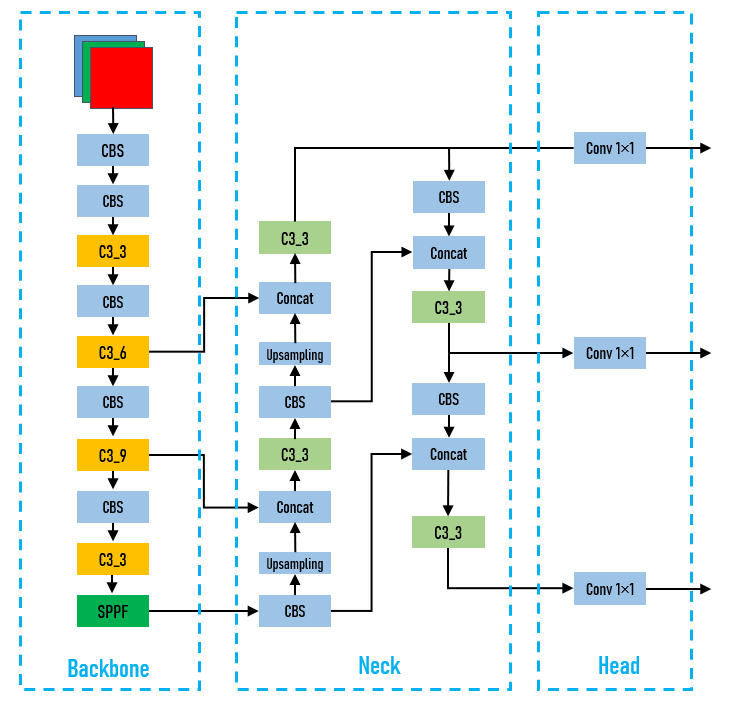}
  \caption{Architecture of YOLOv5. The whole network is composed of Backbone, Neck and Head.
  } \label{fig:yolov5 struc}
\end{figure}

\begin{figure}[htpb]
  \centering
  \includegraphics[width=0.7\textwidth]{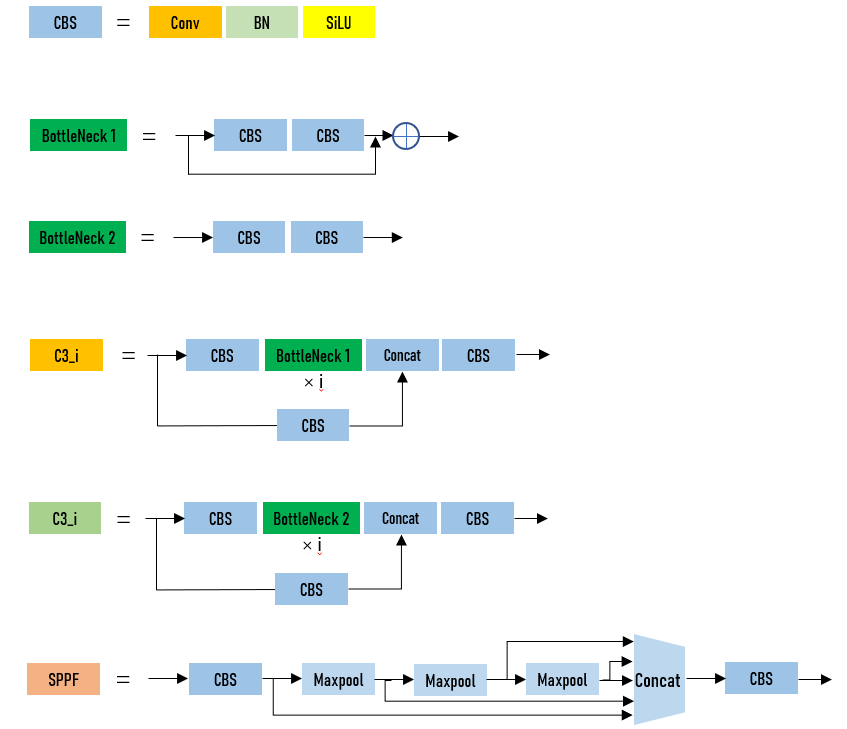}
  \caption{Details of each component in the YOLOv5 backbone.
  } \label{fig:yolov5 component}
\end{figure}

In the previous chapter, we briefly introduced the YOLO family. In this section, we will specifically introduce the latest version of the YOLO algorithm, i.e., YOLOv5 and its network structure. Similar to the previous version of YOLO, the whole YOLOv5 can still be divided into three parts, namely backbone, neck and head, see figure \ref{fig:yolov5 struc} and \ref{fig:yolov5 component}. Backbone can be regarded as the feature extraction network of YOLOv5, and according to its structure and the previous YOLOv4 backbone, we can generally call it CSPDarknet. The input images are first extracted in CSPDarknet, and the extracted features can be called feature maps. In the backbone part, we obtain three feature maps for the next step of network, i.e., the neck part. This part can be also called the enhanced feature extraction network of YOLOv5. The three feature maps obtained in the backbone part are fused in this part, and the purpose of the feature fusion is to combine the feature information from different scales. In the neck part, , the Path Aggregation Network (PAN) structure is used (\cite{liu2018path}), where we not only upsample the features to achieve feature fusion, but also downsample the features again to achieve feature fusion. Head is the classifier and regressor of YOLOv5. With backbone and neck, we have access to three enhanced feature maps. Each feature map has a width, height and number of channels, so we can think of the feature map as a collection of feature pixels, each of which has a number of channels. As in previous versions of YOLO, the detector head in YOLOv5 is composite, i.e., the classification and bounding box regression are implemented by a 1 × 1 convolution. In summary, the entire YOLOv5 network is doing the following: feature extraction - feature enhancement - prediction of objects corresponding to the feature pixels.

As for the CSPDarknet in YOLOv5, it has four important features: (1) Using residual network (\cite{he2016deep}). The residual convolution in CSPDarknet can be divided into two parts, the main part is a 1 × 1 convolution and a 3 × 3 convolution; the skip connection does not do any processing, and directly combining the input and output of the main part. The whole YOLOv5 backbone is composed of residual convolution. The residual structure is characterized by its ease of optimization and its ability to improve accuracy by adding considerable depth. Its internal residual blocks use skip connections to alleviate the problem of gradient vanishing caused by increasing depth in deep neural networks. (2) Using the CSPnet structure (\cite{wang2020cspnet}). The CSPnet structure is not too complicated, which is a splitting of the original stack of residual blocks into two parts: the main part continues the original stack of residual blocks; the other part acts like a skip connection and is directly connected to the end after a small amount of processing. Therefore, it can be considered that there is a large skip connection in the CSP. (3) Using SiLU activation function, which is an improved version of Sigmoid and ReLU. SiLU has the properties of no upper bound with lower bound, smooth, and non-monotonic. SiLU works better than ReLU on deep neural networks and it can be regarded as a smooth ReLU activation function. (4) Using SPPF structure. Feature extraction is performed by maximum pooling with different pooling kernel sizes to improve the receptive field of the network. In YOLOv4, SPP was used inside the neck, and in YOLOv5, the SPP module is used in the backbone.

\subsection{New Head}
\subsubsection{New Head}
\begin{figure}[htpb]
  \centering
  \includegraphics[width=0.7\textwidth]{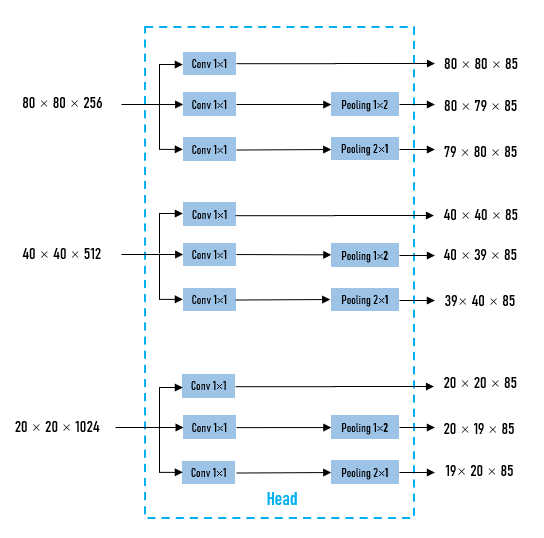}
  \caption{The new head detector outputs 9 feature maps. For the input, it will be divided into three types of processing: Conv, Conv + (1,2) pooling and Conv + (2,1) pooling.
  } \label{fig:new head}
\end{figure}

In the previous section, we explained the structure of YOLOv5, which consists of backbone, neck and head. The first change to YOLOv5 in this thesis is to change the size of the feature map by adding asymmetrical pooling layer to the head part. The feature map sizes are 20 × 20 × 1024, 40 × 40 × 512, and 80 × 80 × 256 before passing through the 1 × 1 convolutional layer, respectively. Take 80 × 80 × 256 feature map as an example, after passing 1 × 1 convolutional layer, its dimension becomes 80 × 80 × 255, i.e., 80 × 80 × 3 (4 + 1 + 80). Number 3 shows that there are three feature maps of dimension 80 × 80. The difference between them is that the pixels in each feature map correspond to different sizes of anchor boxes. For example, the sizes of the three anchor boxes are 10 × 13, 16 × 30, and 33 × 18. The feature points of the three feature maps have the same characteristics, i.e., three groups of feature points have the same size of receptive fields, so they use different anchor boxes to try to fit the different shapes of the object. In our new model, we change both receptive fields and anchor boxes of each feature map. Firstly, the shape of the receptive field corresponding to each group of pixels needs to be changed. I speculate that when the receptive field corresponding to a point is square, the point has better prediction ability for objects whose shape is close to square. When the aspect ratio of the receptive field is 2:1, points can predict objects with the same shape better. The same is true for a receptive field with an aspect ratio of 1:2. So, after the 1 × 1 convolutional layer, I add two asymmetric pooling layers to head part, see figure \ref{fig:new head}. Thus, in our new model, in order to distinguish the role of each anchor box more clearly, one anchor box is used to predict objects close to a square shape, one anchor box is used to predict rectangular objects whose width is larger than their height, and the last anchor box is used to predict rectangular objects whose width is smaller than their height. In summary, the head detector will no longer output 3 feature maps but 9 feature maps. Their sizes are 20 × 20 × 85, 20 × 19 × 85, 19 × 20 × 85, 40 × 40 × 85, 40 × 39 × 85, 39 × 40 × 85, 80 × 80 × 85, 80 × 79 × 85, and 79 × 80 × 85, respectively. For the pooling layer, this thesis chooses average pooling, in order to avoid losing too much context. Since only the pooling layer is added, the number of parameters of the new network does not increase, so it runs just as fast, and at the same time, we think its detection capability will be improved. In fact, one can also try to replace the pooling layer with a convolutional layer with kernel sizes of (1, 2) and (2, 1). Although the network's parameters will increase, its running speed is not significantly affected.

\subsubsection{New Anchors}
Setting anchors in advance is an very important things. In original YOLOv5 model, depending on the size of the feature map, the anchors are divided into three groups: (10,13), (16,30), (33,23) for 80 × 80 feature map , (30,61), (62,45), (59,119) for 40 × 40 feature map and (116,90), (156,198), (373,326) for 20 × 20 feature map. In order to fit our new model better, we consider that the sizes of the anchors also needed to be modified. Therefore, the new anchors are generated. For the nine feature maps in the figure \ref{fig:new head}, the anchors are (20,20), (40,20), (20,40), (60,60), (120,60), (60,120), (400,200), (200,400) in order. The shape of these anchors corresponds to the receptive field of each cell on the 9 feature maps. We assume that the feature map whose cells have rectangular receptive field can detect rectangular object more easily by using rectangular anchor.

\subsubsection{New Strategy of NMS}
As we mentioned before, the head of YOLOv5 model is modified from 3 feature maps to 9 feature maps. Anchors are also modified to adapt to the corresponding feature maps. Finally, we divide 9 feature maps into 3 types: having square receptive fields, having receptive field with an aspect ratio of 2:1, and having receptive field with an aspect ratio of 1:2. We are going to use these three types of feature maps to detect objects with different aspect ratios. In original YOLOv5 model, the NMS is used on the whole predicted boxes and the number of times we use it is one. But in our method, we use four times NMS. NMS is first performed on the boxes predicted by each of the three types of feature maps, and after the results are fused, NMS is finally done again. Using this strategy, we hope that the new model will have better performance for multiple shapes and  categories of objects.

\section{Experiment}
\subsection{Configuration}
The structure of the new model has the same backbone and neck as YOLOv5 (\cite{glenn_jocher_2022_7002879}), only the head is different. Therefore, to ensure the training speed, YOLOv5n is chosen for the new model. We use VS Code as programming platform and the GPU for training and validation is RTX 3080. The machine learning framework is Pytorch. The version of cudatoolkit is 11.3.

\subsection{Training Hyperparameter}
The training parameters need to be adjusted before training, and these parameters are in train.py.

\begin{table}[h]
    \centering
    \begin{tabular}{|c|c|}
    \hline '--weights' & default=' ' \\
    \hline '--cfg' & default='yolov5n.yaml' \\
    \hline '--epochs' & 300 \\
    \hline '--batch-size' & 128 \\
    \hline '--imgsz' & default=640 \\
    \hline ''--noautoanchor'' & default=True \\
    \hline
    \end{tabular}
    \caption{Hyperparameters for training of new model.}
    \label{tab:Hyperparameter}
\end{table}

\subsection{Evaluations}
The goal of this thesis is to improve the YOLOv5 algorithm. We trained the original and modified YOLOv5n models on the COCO dataset and then compare their Precision, Recall and mAP. The trained models are divided into the following categories: (1) original models, i.e., the backbone, neck and head of the model are not changed. (2) Modified models, which can be divided into 4 kinds, contains three square anchors, three 2:1 aspect ratio, three 1:2 aspect ratio anchors, 9 anchors, respectively.

For the evaluation of the models, we similarly divide the process of validation into the following steps: first, we validate the models which only contain anchors with a shape of square, 2:1 aspect ratio and 1:2 aspect ratio, respectively, to obtain three results, and then compare the results of them with original model. It should be noted that the validation sets of different models are different, for example, for the model with only 3 square anchors ([20, 20], [60, 60], [200, 200]), the labels in its validation set are also approximately square. And for the model with anchors with an aspect ratio of 2:1 ([40, 20], [120, 60], [400, 200]), the width of the labels in its validation set is also larger than the height. This is to verify our idea that a square receptive field with a square anchor is better at predicting objects that are approximately square, as well as rectangular receptive field with a rectangular anchor. Secondly, we validate the model having nine feature maps (the anchors are [20, 20], [60, 60], [200, 200], [40, 20], [120, 60], [400, 200], [20, 40], [60, 120], [200, 400], respectively). And the corresponding validation set contains 5000 pictures with complete labels. Finally, we validate the model with original architecture and modified $loss.py$. 

\subsection{Validation on 3-Feature Maps Networks}
\subsubsection{Square-Anchor Model}
\begin{figure}[htpb]
  \centering
  \includegraphics[width=0.9\textwidth]{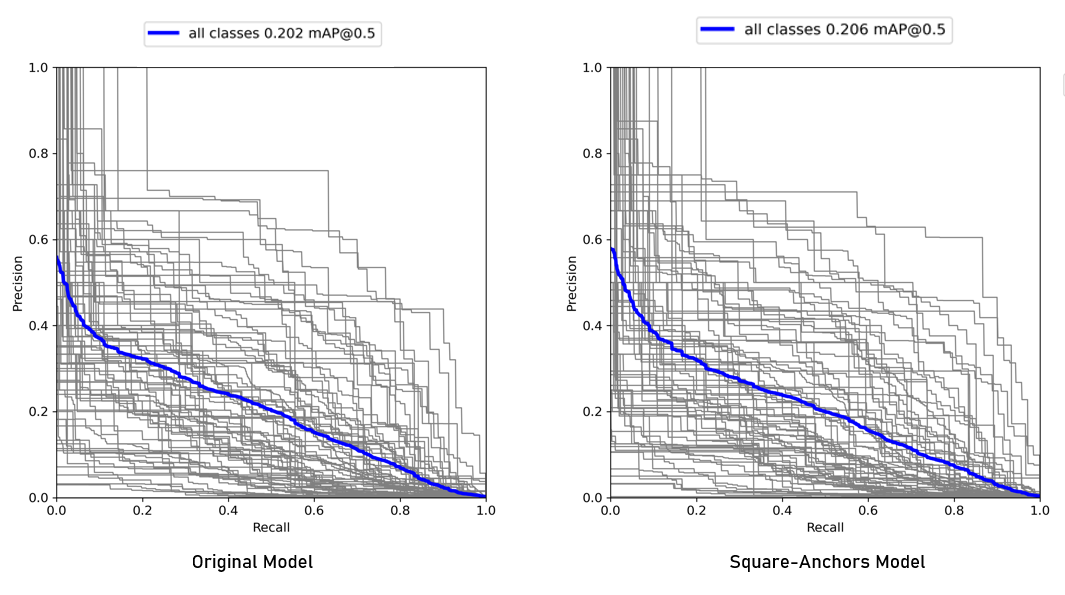}
  \caption{PR-curves of original model and square-anchor model. The mAP@0.5 of the original model is 0.202 and the mAP@0.5 of square-anchors model is 0.206. mAP@0.5 means the threshold of IoU is 0.5.
  } \label{fig:square anchor PR}
\end{figure}

\hspace{1em}In this section, we compares the performance of models with 3 kinds of anchors with the original model. As for the model with 3 square anchors, the validation dataset has 2,988 pictures. Each label in one picture has an aspect ratio between 1/1.2 and 1.2. Figure\ref{fig:square anchor PR} shows the PR-curve of the original model and square-anchor model. The difference between these two model is that the square-anchor model has three feature maps (20 × 20, 40 × 40, 80 × 80), but there is only one square anchor on each feature map. 

\begin{table}[htpb]
    \centering
    \begin{tabular}{|c|c|c|c|c|c|c|c|}
    \hline & P & R & mAP@0.5 & mAP@.5:.95 & pre-process & inference & NMS \\
    \hline Original & 0.243 & 0.39 & 0.202 & 0.127 & 0.7ms & 4.3ms & 2.1ms\\
    \hline Square-anchors & 0.254 & 0.365 & 0.206 & 0.13 & 0.4ms & 4.1ms & 2.3ms \\
    \hline
    \end{tabular}
    \caption{Comparison of original model and square-anchor model. The first four statistics shows the performance of each model. The last three statistics show the processing speed of each image}
    \label{tab:Square anchors}
\end{table}

The result shows that for approximately square labels, the feature maps which have square anchors and square receptive fields have better performance in terms of precision, mAP and processing speed.

\subsubsection{Asymmetrical Average Pooling Model}
\begin{figure}[htpb]
  \centering
  \includegraphics[width=0.9\textwidth]{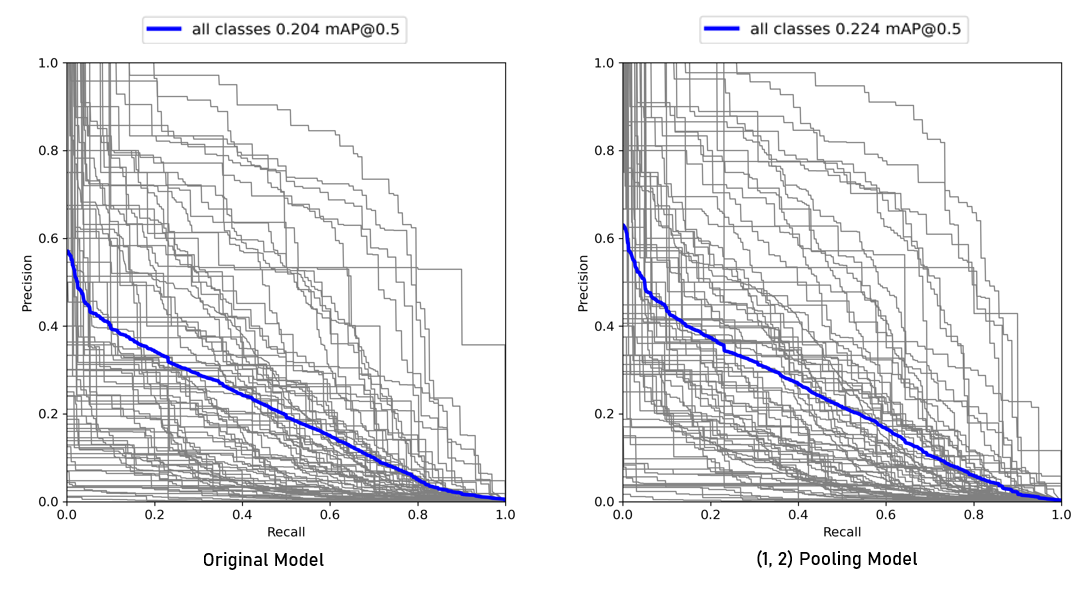}
  \caption{PR-curves of original model and (1, 2) pooling model. The mAP@0.5 of the original model is 0.204 and the mAP@0.5 of square-anchors model is 0.224.
  } \label{fig:21 anchor PR}
\end{figure}

\hspace{1em}We added each of the two asymmetrical average pooling layers to the head of the original network, thus obtaining two new models. For the model, which is added a (1, 2) pooling layer, its aspect ratio of receptive field becomes 2:1. And its validation set has 3,158 images containing 8,522 labels. Each label in one picture has an aspect ratio greater than 1.2. Figure~\ref{fig:21 anchor PR} shows the PR-curve of the original model and (1, 2) pooling model.

\begin{table}[htpb]
    \centering
    \begin{tabular}{|c|c|c|c|c|c|c|c|}
    \hline & P & R & mAP@0.5 & mAP@.5:.95 & pre-process & inference & NMS \\
    \hline Original & 0.267 & 0.356 & 0.204 & 0.118 & 0.7ms & 4.1ms & 2.3ms\\
    \hline (1, 2) pooling & 0.293 & 0.374 & 0.224 & 0.131 & 0.4ms & 4.6ms & 0.9ms \\
    \hline
    \end{tabular}
    \caption{Comparison of original model and (1, 2) pooling model. The first four statistics shows the performance of each model. The last three statistics show the processing speed of each image}
    \label{tab:21 anchors}
\end{table}

\begin{figure}[htpb]
  \centering
  \includegraphics[width=0.9\textwidth]{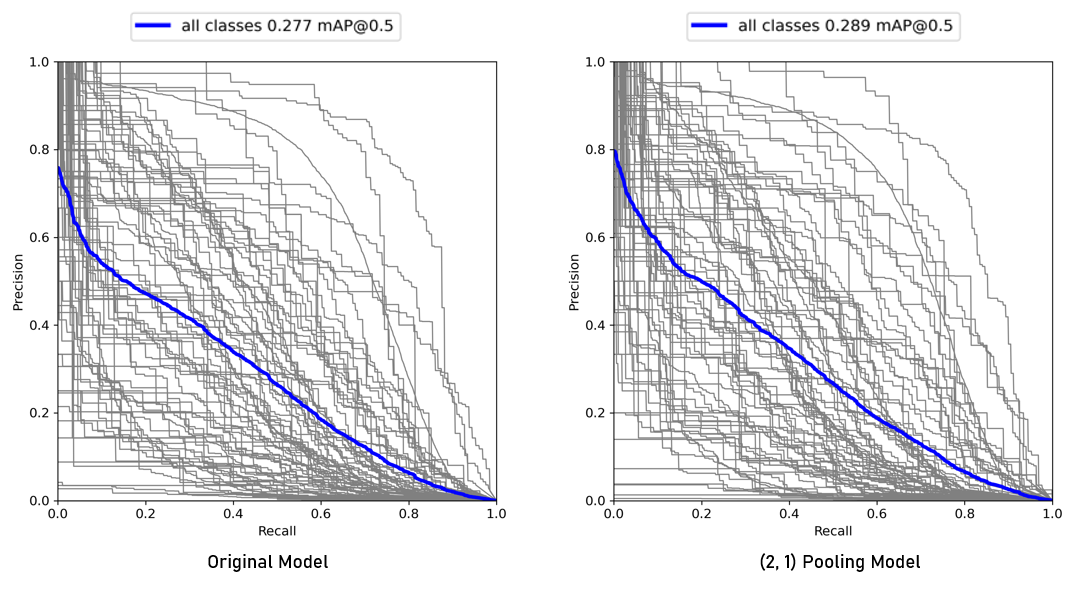}
  \caption{PR-curves of original model and (2, 1) pooling model. The mAP@0.5 of the original model is 0.227 and the mAP@0.5 of (2, 1) pooling model is 0.289.
  } \label{fig:12 anchor PR}
\end{figure}

The result shows that for the labels, whose width is greater than height, the feature maps which have rectangular anchors and rectangular receptive fields have better performance in terms of precision, mAP and processing speed. And for the model, which is added a (2, 1) pooling layer, its aspect ratio of receptive field becomes 1:2. And its validation set has 4,061 images containing 21,578 labels. Each label in one picture has an aspect ratio less than 1/1.2. Figure~\ref{fig:12 anchor PR} shows the PR-curve of the original model and (2, 1) pooling model.

\begin{table}[htpb]
    \centering
    \begin{tabular}{|c|c|c|c|c|c|c|c|}
    \hline & P & R & mAP@0.5 & mAP@.5:.95 & pre-process & inference & NMS \\
    \hline Original & 0.364 & 0.386 & 0.277 & 0.171 & 0.7ms & 4.5ms & 1.8ms\\
    \hline (2, 1) pooling & 0.417 & 0.363 & 0.289 & 0.172 & 0.7ms & 4.1ms & 0.9ms \\
    \hline
    \end{tabular}
    \caption{Comparison of original model and (2, 1) pooling model. The first four statistics shows the performance of each model. The last three statistics show the processing speed of each image}
    \label{tab:12 anchors}
\end{table}

For the labels, whose height is greater than width, the feature maps which have rectangular anchors and rectangular receptive fields also have better performance in terms of precision, mAP and processing speed. 

\subsection{Validation on 9-Feature Maps Network}
\hspace{1em}From previous section, we find that all three new networks perform better than original network in specific validation sets. Now, we combine these three 3-feature maps networks to get a 9-feature maps network, as shown in figure~\ref{fig:new head}. And the validation dataset contains complete 5,000 images, 36,335 labels.

\begin{figure}[htpb]
  \centering
  \includegraphics[width=0.9\textwidth]{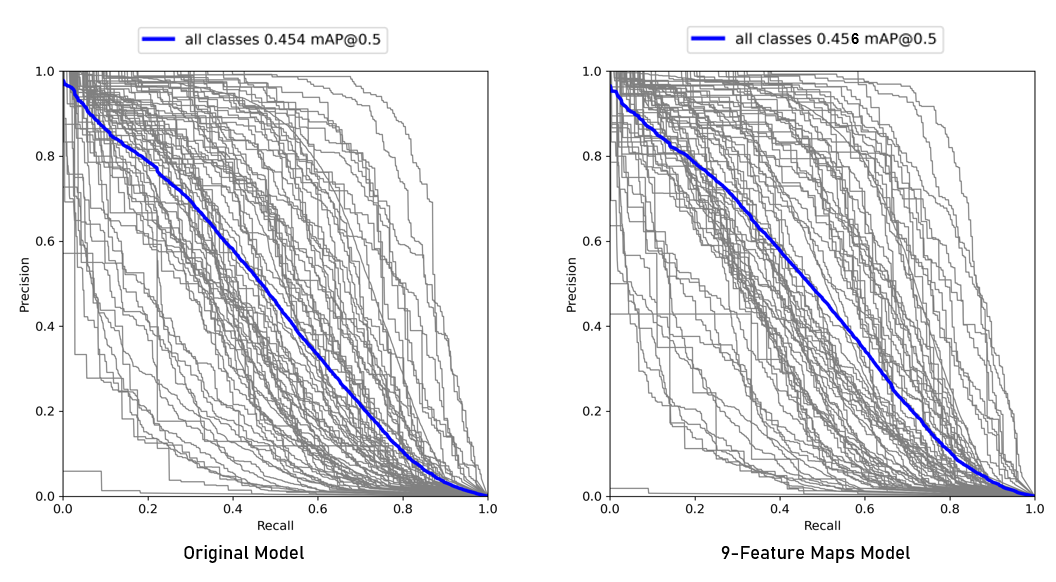}
  \caption{PR-curves of original model and 9-feature maps model. The mAP@0.5 of the original model is 0.454 and the mAP@0.5 of 9-feature maps model is 0.456.
  } \label{fig:9 feature maps PR}
\end{figure}

\begin{table}[htpb]
    \centering
    \begin{tabular}{|c|c|c|c|c|c|c|c|}
    \hline & P & R & mAP@0.5 & mAP@.5:.95 & pre-process & inference & NMS \\
    \hline Original & 0.597 & 0.418 & 0.454 & 0.267 & 0.6ms & 4.7ms & 1.7ms\\
    \hline 9-Feature Maps & 0.596 & 0.427 & 0.456 & 0.269 & 0.6ms & 5.0ms & 2.6ms \\
    \hline
    \end{tabular}
    \caption{Comparison of original model and (2, 1) pooling model. The first four statistics shows the performance of each model. The last three statistics show the processing speed of each image}
    \label{tab:9-Feature Maps}
\end{table}

Figure~\ref{tab:9-Feature Maps} shows that the 9-feature maps model performs better than original model in terms of recall and mAP, but the speed is slower. This is because the input of the image need to be processed in 9 feature maps instead of 3. Meanwhile, extra pooling layers also increase the processing time. And in NMS step, we do 4 times NMS.

\section{Conclusion and Future Work}
Object detection has been a hot topic in recent years, and with the continuous efforts of researchers, object detection algorithms are performing better and better. Their accuracy and speed are also gradually able to meet the needs of various industries. For example, autonomous driving is also a booming industry, and its high requirements for object detection have promoted further improvements in the effectiveness of object detection algorithms.

This article is based on the original YOLOv5 with some modifications. As a result, the accuracy of the algorithm is improved while ensuring the speed. Specifically, the backbone and neck parts of the new network are the same as the original one, because the facts tell us that they perform well enough. We finally chose to change the head part. The output of the model, i.e., three square feature maps, was changed to nine, and six of them are no longer square. The previous layer of these six feature maps was a newly added asymmetrical pooling layer, so that we can change the receptive fields of the feature maps without adding a new number of parameters, to expect the model to have better predictive power for multiple shapes of objects. 
The final experimental results show that the new model is indeed improved. Its mAP is improved by 0.002 compared to the original model, but its inference speed is not affected too much. Compared with the original YOLOv5 model, the new model has advantages in terms of detection accuracy. In the future, firstly, we can continue to optimize the network structure and further improve the accuracy. In addition to modifying the head part, we can also try to modify other structures, for example, backbone and neck. Secondly, the prediction speed of the model has further room for improvement. Finally, the model can be applied to autonomous driving. For example, in an autonomous driving simulation system.


\bibliographystyle{unsrt} 
\bibliography{test}

\end{document}